\def\year{2019}\relax
\documentclass[letterpaper]{article} 
\usepackage{aaai19}  
\usepackage{times}  
\usepackage{helvet}  
\usepackage{courier}  
\usepackage{url}  
\usepackage{graphicx}  
\usepackage{units}
\usepackage{colortbl}
\definecolor{Gray}{gray}{0.8}
\usepackage{amsthm,amsmath}
\usepackage[inline]{enumitem}
\usepackage{tikz}
\usepackage[framemethod=tikz]{mdframed}
\usetikzlibrary{arrows,plotmarks,decorations.markings,trees,shapes}
\tikzstyle{nodo}=[ellipse,draw=black!100,fill=black!30,line width=.7pt,minimum width=1.2cm,minimum height=.7cm]
\tikzstyle{Qnodo}=[ellipse,draw=black!100,fill=black!10,line width=.7pt,minimum width=1.2cm,minimum height=.7cm]
\tikzstyle{arco}=[draw=black!80,line width=.7pt, postaction={decorate}, decoration={markings,mark=at position 1.0 with {\arrow[ draw=black!80,line width=.7pt]{>}}}]
\tikzstyle{decision} = [rectangle, draw, fill=black!100,text=white, text width=4.5em, text badly centered, node distance=3cm, minimum height=3em]
\tikzstyle{block} = [rectangle, draw, fill=blue!20, text width=5em, text centered, rounded corners, minimum height=3em]
\tikzstyle{line} = [draw, -latex']
\tikzstyle{cloud} = [draw, ellipse,fill=red!20, node distance=3cm, minimum height=2em]
\theoremstyle{definition}
\newtheorem{Example}{Example}[]

\begin{document}
\title{A Bayesian Approach to Conversational Recommendation Systems}
\author{F. Mangili$^*$, D. Broggini$^*$, A. Antonucci$^*$, M. Alberti$^\dagger$ \and L. Cimasoni$^\dagger$\\ $^*$IDSIA (Switzerland) - {\tt idsia.ch} \\
$^\dagger$Stagend (Switzerland) - {\tt stagend.com}}
\maketitle

\begin{abstract}
We present a conversational recommendation system based on a Bayesian approach. A probability mass function over the items is updated after any interaction with the user, with information-theoretic criteria optimally shaping the interaction and deciding when the conversation should be terminated and the most probable item consequently recommended. Dedicated elicitation techniques for the prior probabilities of the parameters modeling the interactions are derived from basic structural judgements. Such prior information can be combined with historical data to discriminate items with different recommendation histories. A case study based on the application of this approach to \emph{stagend.com}, an online platform for booking entertainers, is finally discussed together with an empirical analysis showing the advantages in terms of recommendation quality and efficiency.
\end{abstract}

\section{Introduction}
The task of selecting from a collection of items one which is in some sense optimal for a specific user is a classical AI task. Several algorithms have been explored to perform such tasks and automated recommendations are today included in most modern e-commerce websites \cite{he2016interactive,sarwar2000analysis}. In standard setups, no interaction with users is considered, and the recommendation system bases its decision on some background information about the user, historical records of her choices and of the choices of other similar users, and, more recently, automatically inferred contextual information \cite{adomavicius2011context,lu2015recommender}.

Yet, modern technologies such as chat-bots or personal assistants need systems able to support and model dynamic and sequential interactions, this leading to a substantial re-design of the traditional approaches. Here we focus on such a newer class of recommendation systems, called here \emph{conversational}, as we term conversation a sequence of dynamically customized interactions between the user and the system, before the latter return the recommendation. This class of systems exploits the knowledge-based recommendation techniques and is based on a strong interaction with the users. Therefore, this type of recommendation system should be considered when the goal is to support the user in purchasing a high-involvement product. In such a situation, indeed, the user wants to be involved in the decision process, and thus is not bothered by the need of interacting with the system \cite{jugovac2017interacting}.

To achieve that, we take inspiration from existing approaches in the field of computer testing \cite{wainer2000computerized,butz2006web,mangili2017reliable}, whose goal is to determine the skills of a student on the basis of the answers to the questions of a test. In particular we focus on \emph{adaptive} approaches in which the system selects the next question to ask to the student from a given set of questions on the basis of the previous answers by information-theoretic scores, that are also used to decide when the test should be ended. This can be easily achieved with generative probabilistic models such as, for instance, Bayesian networks \cite{Millan2002}, which are sequentially updated any time the answer to a new question is collected. The adaptive concept is converted here in a conversational approach which will lends itself to the (future) development of a dynamic generation of questions and richer interaction models.

The conversational procedure is applied to the recommendation system of Stagend (stagend.com), an online platform that allows organizers of events (playing here the role of the users) to book artists (playing here the role of the items) for their events. The goal of the recommendation system is therefore to select the most suitable artist for a particular event based on a set of needs elicited by the user. Results of some simulations based on data from Stagend platform will be presented to show the advantages provided by the conversational approach compared to earlier procedure based on a static questionnaire and discuss possible improvements.


\section{Basic Concepts}
Consider a recommendation system based on a \emph{catalogue} $\mathcal{I}$ including $n$ items, say $\mathcal{I}:=\{i_1,\ldots,i_n\}$. The system is supposed to support the user in selecting a single item from the catalogue on the basis of a conversational process. We assume that, at the end of the conversation, the user always picks one item from $\mathcal{I}$. The uncertain quantity $I$ denotes the element of $\mathcal{I}$ to be picked by the user. We consider a Bayesian setup and model the subjective probabilities of picking the different items before the start of the conversational process as a prior mass function $P(I)$.

We call \emph{questions}, the interactions between the user and the system. A generic question is denoted as an uncertain quantity $Q$ taking its possible values, to be called \emph{answers}, from $\mathcal{Q}$. Here we assume a finite set of possible answers for each question, say $\mathcal{Q}:=\{q^i,\ldots,q^r\}$. As a model of the relation between $Q$ and $I$, we might be able to assess a \emph{conditional probability table} $P(Q|I)$ whose columns are indexed by the answers and whose rows are associated to the items. After an answer $Q=q \in \mathcal{Q}$ is collected, the probability mass function over the items can be updated by Bayes theorem, i.e.,
\begin{equation}\label{eq:bayes}
P(i|q) = \frac{P(i) \cdot P(q|i)}{\sum_i P(i) \cdot P(q|i)}\,.
\end{equation}
This shows that the impact of an answer $Q=q$ to the choice $I$ is only based on the relative proportions of the values of $P(q|i)$ for each $i\in\mathcal{I}$. In particular, setting an element of the conditional probability table equal to zero implies a logical constraint for which the answer associated with column $Q=q$ makes the choice of the item associated with row $I=i$ impossible.

On the other side, decisions of the recommendation system are based on the the conditional probability $P(i|q)$ which may be strongly influenced by the prior $P(i)$. Therefore, to better elicit the behaviour of the model, understand its underpinning assumptions and foresee the system behaviors, it will be often more natural to consider the joint probability $P(i,q)$, rather than the conditional one $P(q|i)$. In fact, as the  probabilities of the joint events $(i,q)$ completely describe the model, they allow on one hand to derive the probabilities of interest $P(q|i)$ and $P(i)$ and, on the other hand,  to reason about the relative weight of the posterior probabilities assigned to different items after conditioning on the answer, as, for strictly positive probabilities,  it holds:
\begin{equation}
\frac{P(i',q)}{P(i'',q)} = \frac{P(i'|q)}{P(i''|q)}\,,
\end{equation}
for each $i',i''\in\mathcal{I}$ and $q\in\mathcal{Q}$. 
 
We start by considering a static approach to the elicitation of the user needs, based on a list of $m$ questions $\mathbf{Q} = \{Q_1,\dots, Q_m\}$, called here a \emph{questionnaire}. Selection of the optimal item to be suggested to a user at the end of the conversation is based on the conditional probability $P(i|\mathbf{q})$ of each item $i$ given the list of answers collected, hereafter described by the vector $\mathbf{q}$. In order to improve the user experience it is desirable to minimize the number of questions necessary to identify the optimal suggestion. To this goal, the list of questions in  $\mathbf{Q}$ should be built dynamically as the conversation proceeds. Such a customized list of questions is called hereafter a \emph{conversation}. Before discussing this dynamic approach, we will present the set up that allows the computation of $P(i|\mathbf{q})$, be $\mathbf{q}$ the output of a questionnaire or of a conversation. To model a conversation we initially formulate a naive-like assumption stating the conditional independence of the questions given the item, i.e.,
\begin{equation}\label{eq:indep}
P(q_1,\ldots,q_m|i) = \prod_{j=1}^m P(q_j|i)\,.
\end{equation}
More general models will be discussed later on in this paper. 
Under this assumption, if the conditional probability tables $P(Q|I)$ associated to each $Q\in\mathbf{Q}$ are available, the probability $P(i|\mathbf{q})$ can be obtained by recursively applying the Bayes theorem to update the probability $P(i|q_1,\ldots,q_k)$ after the first $k$ answers by conditioning also on the next answer $q_{k+1}$.

To open the discussion, let us present a toy example, inspired by the Stagend case study reported in the final part. The example will be used through the paper to illustrate the main features of the proposed method.
\begin{Example}\label{ex:1}
{\it Stagend users are invited to answer a number of questions about the entertainment they are looking for. The two questions $Q_1$ and $Q_2$ typically asked to the users are reported in Table \ref{tab:questions}, while a catalogue of three entertainers-items is in Table \ref{tab:catalogue}.}
\end{Example}

\begin{table}[htp!]
    \centering
    \begin{tabular}{cp{6cm}}
        \hline
        $\mathcal{I}$&Catalogue\\
        \hline
        $i_1$&DJ available for all type of events\\
        $i_2$&Band available for weddings and corporate events\\
        $i_3$&Magician available for birthdays and parties for kids\\
        \hline
    \end{tabular}\caption{A catalogue}\label{tab:catalogue}
\end{table}

\begin{table}[htp!]
\centering
\begin{tabular}{cp{6cm}}
\hline
$Q_1$&Which entertainment are you looking for?\\
\hline
$q_1^1$ & DJ\\
$q_1^2$ & Band\\
$q_1^3$ & Musician\\
$q_1^4$ & Entertainer\\
\hline
\end{tabular}
\vskip 2mm
\begin{tabular}{cp{6cm}}
\hline
$Q_2$&Which event are you organizing?\\
\hline
$q_2^1$ & Wedding\\
$q_2^2$ & Corporate event\\
$q_2^3$ & Birthday\\
$q_2^4$ & Party for kids\\
\hline
\end{tabular}
\caption{Two questions of Stagend questionnaire}\label{tab:questions}
\end{table}

\section{Elicitation by Logical Compatibility}
Consider the background information about the entertainers in Table \ref{tab:catalogue} and the questions (and possible answers) in Table \ref{tab:questions}. An underlying notion of \emph{logical compatibility} can be therefore formalized as follows. Given an item $i\in\mathcal{I}$ and an answer $q$ to a question $Q\in\mathbf{Q}$, we say that $i$ is compatible with  $q$, and we denote this as $i\models q$, if $i$ satisfy all the logical requirements implied by $q$. As an example $i_2 \models q_2^2$, i.e., the second antertainer in the catalog is compatible with a corporate event, but $i_3 \not\models q_i^3$, i.e., the third entertainer is not a musician. Given $Q\in\mathbf{Q}$, we can therefore define an indicator function for such a compatibility concept for each pair  $q\in\mathcal{Q}$ and $i\in\mathcal{I}$ as follows:
\begin{equation}
\delta(i,q) :=
\left\{
\begin{array}{ll}
1 & \text{if} \quad i\models q\\
0 & \text{otherwise}\,.
\end{array}
\right.
\end{equation}

Following the above definition of compatibility, each item can be characterized by the set of compatible answers for each of the $m$ possible questions. For instance, artist $i_2$ in Example \ref{ex:1} is described by set of answers $\{q_1^2\}$ to $Q_1$ and   $\{q_2^1,q_2^2\}$  to $Q_2$.

In the probabilistic model proposed, the notion of compatibility is translated into the assumption that the joint probability $P(q,i)$ and, following Equation \eqref{eq:bayes}, also the conditional probability $P(q|i)$, have the same support of  $\delta(i,q)$, i.e., they are zero whenever $i \not\models q$. To elicit $P(q,i)$, and hence $P(q|i)$, when $q$ and $i$ are compatible, we need additional sets of assumption. In the remaining part of this section we discuss strategies to tackle this problem and analyze the underlying assumptions and consequences in terms of posterior inferences.

\subsection{Items Compatible with a Single Answer}
The above discussed notion of logical compatibility might be sufficient to elicit a conditional probability table $P(Q|I)$ without any additional assumption in the special case in which each item is compatible only with a single answer. Let $q^*(i)$ denote such answer in the case of item $i$, i.e., $i \models q^*(i)$ and $i \not \models q$ for each $q\neq q^*(i)$. Under these assumptions, the number of non-zero elements of the joint mass function $P(Q,I)$ is equal to the catalogue cardinality $k:=|\mathcal{I}|$. We assume all these non-zero probabilities to be equal, this meaning $P(q^*(i),i) = k^{-1}$ for each $i\in\mathcal{I}$, while the corresponding conditional probabilities $P(q^*(i)|i)$ have value one (e.g., see Table \ref{tab:conditional1}). After the conditioning on the observed answer $q$, all the items compatible with $q$ will have the same posterior probability, whereas all non compatible items will receive zero probability mass. Therefore, in this simple setting, the Bayesian approach proposed implement a logical filter, making impossible the items incompatible with the answers.
Such a situation occurs in the following example.

\begin{Example}\label{ex:2}
{\it In the setup of Example \ref{ex:1}, consider the conditional probability table for $Q_1$. Each entertainer is consistent with only one a single answer to this question. The corresponding quantification of the conditional probability table is depicted in Table \ref{tab:conditional1}. If we assume, for instance, that a new user answer 1 to the first question $Q_1$, we can update the probabilities assigned to each entertainer by Equation \eqref{eq:bayes} and conclude a posterior probability equal to one for artists $i_1$ and zero for the other two entertainers.}
\end{Example}

\begin{table}[htp!]
\centering
\begin{tabular}{ccccc}
\hline
$P(q_1|i)$&$q_1^1$&$q_1^2$&$q_1^3$&$q_1^4$\\
\hline
$i_1$&$1$&$0$&$0$&$0$\\
$i_2$&$0$&$1$&$0$&$0$\\
$i_3$&$0$&$0$&$0$&$1$\\
\hline
\end{tabular}
\caption{Elicitation of $P(Q_1|I)$}\label{tab:conditional1}
\end{table}

\subsection{Items Compatible with Multiple Answers}
Consider a question $Q\in\mathbf{Q}$ such that the assumptions in the previous section are not satisfied and there are items compatible with more than a single answer. In this case, the elicitation of the joint mass function $P(Q,I)$ requires further assumptions. Here below we describe two alternative strategies for such elicitation. It is a trivial exercise to note that both strategies lead to the elicitation discussed in the previous case for items compatible with a single answer, that is a special case of the one considered here.

\paragraph{Uniform Joint Strategy (UJS).} Let us assume, as in the previous section, that all the compatible item/question combinations, i.e., all the pairs $(i,q)$, with $i\in\mathcal{I}$ and $q\in\mathcal{Q}$, such that $i\models q$ have the same probability. If $N(Q)$ denotes the number of these pairs, this assumption corresponds to assigning  $P(i,q):=N(Q)^{-1}$. Let us also define the \emph{versatility} $v_Q(i)$ of an item $i\in\mathcal{I}$ with respect to a question $Q$ as the number of answers of $Q$ that are compatible with the item $i$.  Then, $N(Q)=\sum_{i\in \mathcal{I}} v_Q(i)$. From the joint mass function $P(I,Q)$, we obtain $P(i)=v_Q(i)/N(Q)$ and $P(q|i)=\delta(i,q)/v_Q(i)$ for each $i\in\mathcal{I}$. 

It is easy to note that, before the question is asked, the more versatile an item is, the higher is its probability of being selected. However, once the desired characteristic of the item has been elicited by the user, all item compatible with it will be assigned the same posterior probability. 

\begin{Example}\label{ex:ujs}
{\it In the setup of Example \ref{ex:1}, UJS can be applied to the elicitation of the conditional probability table for question $Q_2$. This corresponds to the joint probability mass function in Table \ref{tab:joint2}. By summing over the rows we obtain the corresponding marginal probabilities for the entertainers: $P(i_1)=\nicefrac{1}{2}$, and $P(i_2)=P(i_3)=\nicefrac{1}{4}$, while the corresponding conditional probability table is the one depicted in Table \ref{tab:conditional2}. Note that, before the question, entertainer $i_1$ has twice the probability of the other to be chosen, as it is the most versatile one. However, once we collect the answer, e.g., $Q_2=q_2^1$ the two compatible entertainers, i.e, $i_1$ and $i_2$, receive the same conditional probability:
\begin{eqnarray}
P(i_1|q_2^1) \propto P(q_2^1|i_1)\cdot P(i_1)&=& \nicefrac{1}{8}\,,\\
P(i_2|q_2^1) \propto P(q_2^1|i_2)\cdot P(i_2)&=& \nicefrac{1}{8}\,.
\end{eqnarray}}
\end{Example}

\begin{table}[htp] 
\centering
\begin{tabular}{ccccc}
\hline
$P(i,q_2)$&$q_2^1$&$q_2^2$&$q_2^3$&$q_2^4$\\
\hline
$i_1$&$\nicefrac{1}{8}$&$\nicefrac{1}{8}$&$\nicefrac{1}{8}$&$\nicefrac{1}{8}$\\
$i_2$&$\nicefrac{1}{8}$&$\nicefrac{1}{8}$&$0$&$0$\\
$i_3$&$0$&$0$&$\nicefrac{1}{8}$&$\nicefrac{1}{8}$\\
\hline
\end{tabular}
\caption{UJS elicitation of $P(Q_2,I)$}\label{tab:joint2}
\end{table}

\begin{table}[htp]
\centering
\begin{tabular}{ccccc}
\hline
$P(q_2|i)$&$q_2^1$&$q_2^2$&$q_2^3$&$q_2^4$\\
\hline
$i_1$&$\nicefrac{1}{4}$&$\nicefrac{1}{4}$&$\nicefrac{1}{4}$&$\nicefrac{1}{4}$\\
$i_2$&$\nicefrac{1}{2}$&$\nicefrac{1}{2}$&$0$&$0$\\
$i_3$&$0$&$0$&$\nicefrac{1}{2}$&$\nicefrac{1}{2}$\\
\hline
\end{tabular}
\caption{UJS elicitation of $P(Q_2|I)$}\label{tab:conditional2}
\end{table}

\paragraph{Uniform Prior Strategy (UPS).}
As shown by Example \ref{ex:ujs}, UJS might produce non-uniform values for the marginal probability mass function over the items. Alternatively, we can impose such a ``prior'' uniformity, i.e., $P(i)=n^{-1}$ for each $i\in\mathcal{I}$. If we also assume that, given the item, the probability mass is uniformly distributed over all the compatible answers, we have $P(i,q)=(n\cdot v_Q(i))^{-1}$ and, for the conditional probability, $P(q|i) =v_Q(i)^{-1}$. A consequence of this model is that compatible items may have different posterior probabilities, with the more versatile ones being less probable than the less versatile ones. Such a model can then be more suitable to situations where specialization of the item can be considered a positive feature.

\begin{Example}\label{ex:model2}
{\it Here, still in the setup of Example 1, we apply UPS to the elicitation of the conditional probability table for question $Q_2$. By definition, the prior is uniform, i.e., $P(i) = \nicefrac{1}{3}$ for all entertainers, and the conditional probability table is the same as before. In this case, all items have the same prior probability, whereas, after collecting answer $Q_2=q_2^1$, being less versatile, user $i_2$ is assigned a larger probability than $i_1$, coherently with the fact that the joint distribution assigned a larger probability with the joint event $(i_2,q_2^1)$ than to the event $(i_1, q_2^1)$:
\begin{eqnarray}
P(i_1|q_2^1) \propto P(q_2^1|i_1)\cdot P(i_1) = \frac{1}{12}\,,&\\
P(i_2|q_2^1) \propto P(q_2^1|i_2)\cdot P(i_2) = \frac{1}{6}\,.&
\end{eqnarray}}
\end{Example}

\begin{table}[htp] 
\centering
\begin{tabular}{ccccc}
\hline
$P(i,q)$ & $q_2^1$&$q_2^2$&$q_2^3$&$q_2^4$\\
\hline
$i_1$ & $\nicefrac{1}{12}$ & $\nicefrac{1}{12}$& $\nicefrac{1}{12}$& $\nicefrac{1}{12}$\\
$i_2$& $\nicefrac{1}{6}$ & $\nicefrac{1}{6}$ &$0$&$0$\\
$i_3$&$0$&$0$&$\nicefrac{1}{6}$ & $\nicefrac{1}{6}$\\
\hline
\end{tabular}
\caption{UPS elicitation of $P(Q_2,I)$}
\end{table}

\subsection{Coping with Multiple Questions}
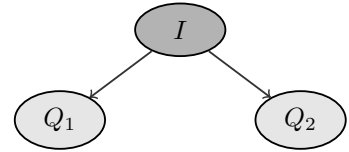
\begin{figure}[htp] 
\centering
\begin{tikzpicture}[scale=.8]
\node[nodo] (i)  at (3,0) {$I$};
\node[Qnodo] (q1)  at (1.,-1.5) {$Q_1$};
\node[Qnodo] (q2)  at (5,-1.5) {$Q_2$};
\draw[arco] (i) -- (q1);
\draw[arco] (i) -- (q2);
\end{tikzpicture}
\caption{Two questions about the item}
\label{fig:bn1}
\end{figure}

In general, the complete list of $m$ possible questions can be divided into $r_{\mathrm{J}}$ questions $\mathbf{Q}^{\mathrm{J}}$ for which we assume UJS, and a set or $r_{\mathrm{P}}$ questions $\mathbf{Q}^{\mathrm{P}}$ for which we assume UPS (if the single answer compatibility holds for a question, this can be can be arbitrarily included in $\mathbf{Q}^{\mathrm{P}}$ or $\mathbf{Q}^{\mathrm{J}}$). 
In such general case, because of Equation \eqref{eq:indep}, the joint mass function is given by:
\begin{equation} \label{eq:genjoin}
P(i,\mathbf{q}) = P(i) P(\mathbf{q}^{\mathrm{J}}|i) P(\mathbf{q}^{\mathrm{P}}|i)\,,
\end{equation}
where, coherently with the UJS and the UPS strategies described above, we have set
\begin{equation} 
P(\mathbf{q}^{\mathrm{P}}|i) = 
\left\{
\begin{array}{ll}
1 & \text{if } r_P = 0\,,\\
\prod_{Q_j \in \mathbf{Q}^{P}} \frac{\delta(i,q_j)}{v_{Q_j}(i)} & \text{otherwise}\,,
\end{array}
\right.
\end{equation}
and
\begin{equation} \label{eq:pipqi}
 P(i) P(\mathbf{q}^{\mathrm{J}}|i)=
\left\{
\begin{array}{ll}
n^{-1} & \text{if } r_J = 0\,,\\
\prod_{Q_j \in \mathbf{Q}^{J}} \frac{\delta(i,q_j)}{N(Q_j)} & \text{otherwise}\,,
\end{array}
\right.
\end{equation}
from which it follows that $P(q|i) = \delta(i,q_j) [v_{Q_j}(i)]^{-1}$ for all $q \in  \mathbf{Q}^{\mathrm{J}}$ and $\mathbf{Q}^{\mathrm{P}}$, and 
\begin{equation} 
P(i) =
\left\{
\begin{array}{ll}
n^{-1} & \text{if } r_J = 0\,,\\
\prod_{Q_j \in \mathbf{Q}^{J}} \frac{v_{Q_j}(i)}{N(Q_j)} & \text{otherwise}\,.
\end{array}
\right.
\end{equation}
\begin{Example}\label{ex:generalmodel}
{\it In the Stagend example the complete model can be graphically represented by Figure \ref{fig:bn1}, where nodes represent variables, edges describe relations of conditional dependence and the absence of a path connecting two nodes corresponds to conditional independence of the corresponding variables. The conditional probability tables  for $P(Q_1|I)$ and $P(Q_2|I)$ are the ones in Tables \ref{tab:conditional1} and \ref{tab:conditional2}, respectively, whereas the prior $P(I)$ is given in Table \ref{tb:prior} and depends on the strategy adopted for modeling question $Q_2$.}
\end{Example}

\begin{table}[htp]
\centering
\begin{tabular}{lccc}
\hline
Strategy&$P(i_1)$&$P(i_2)$&$P(i_3)$\\
\hline
UJS&$\nicefrac{1}{2}$&$\nicefrac{1}{4}$&$\nicefrac{1}{4}$\\
UPS&$\nicefrac{1}{3}$&$\nicefrac{1}{3}$&$\nicefrac{1}{3}$\\
\hline
\end{tabular}
\caption{UJS and UPS elicitations of $P(I)$}
\label{tb:prior}
\end{table}

\section{Adaptive Approach}
\paragraph{Shaping the Conversation.} In classical recommendation systems the assessment of the user preferences with respect to the different items is based on a static block of background information about the user. Such information can be already available in the system or directly obtained from the user after some kind of reduced interaction, e.g., a predefined questionnaire. However, as discussed in the introduction, in modern setups the process of collecting information about the user preferences with respect to the catalogue should be based on a sequence of dynamic interactions. 
In this view, the questionnaire approach leaves the place to a conversational process taking the form of a personalized sequence of questions dynamically picked from a larger set of questions. 
The prior probability mass function $P(I)$ is thus sequentially updated each time a new answer is collected, and the updated probability $P(I|\mathbf{Q})$ is used to select the most informative next question. 
 The choice between a possibly huge set of candidate question/interaction can be driven by information-theoretic criteria making any sequence potentially different from the other. In particular, taking inspiration from the literature in the field of adaptive testing, we pick the question that minimizes the conditional entropy (and hence maximizes the expected information gain). This choice is the most natural one as it allows reducing the entropy of the mass function $P(I|\mathbf{Q})$ in order to concentrate most of the probability mass on a limited number of items.
More formally, the adaptive conversational process selects the question $Q_j^*$ such that:
\begin{equation}
j^* = \arg \min_{j=1,\ldots,m} H(I|Q_j)\,,
\end{equation}
where $H(I|Q):=\sum_{q\in\mathcal{Q}} H(I|q) P(q)$ and $H(I|q)$ is the entropy of the posterior mass function over $I$ after the answer $q\in\mathcal{Q}$ to the question $Q$, and $\{Q_j\}_{j=1}^m$ is the set of questions the system can choose from. 

\paragraph{Stopping Rule.} This procedure is iterated after any answer and the conversation ends if the posterior entropy $H(I|q)$ decreases under a fixed threshold. As that the entropy of a mass function $P(I)$ is defined $H(I):=-\sum_{i\in \mathcal{I}} P(i) \log_{|\mathcal{I}|} P(i)$, a natural threshold $H^*_s$ is the entropy of a mass function over $I$ which is uniform on $s$ items, while the other ones have zero probability, i.e., $H^*_s:=-\log_{|\mathcal{I}|}s^{-1}$. Setting this value in the stopping rule forces the system to halt when most of the posterior probability mass is concentrated on the $s$ most probable items.

\section{Adding Properties}
The approach discussed so far allows to properly model the relation between items and questions and dynamically update the probabilities during the conversation. Yet, such an approach is suitable only for small catalogues, as it implicitly requires the assessment of the logical compatibility relations between each item and each questions. As an example, in the case study under consideration, the catalogue includes more than three thousand items (i.e., entertainers), this preventing a straightforward application of the above outline ideas. To bypass such a limitation, we have already introduced a characterization of each item based on the set of compatible answers to each questions. Here, we formalize this characterization and make it independent from the set of questions by introducing the concept of item \emph{properties}. A property $C$ is a random quantity taking its values in a finite set $\mathcal{C}:=\{c_1,\dots,c_r\}$. Let $\mathbf{C}:=(C_1, C_2,\dots,C_p)$ denote the joint set of relevant properties used to characterize an item
We assume that this set of properties is a sufficient description of the item $I$ and, consequently, the questions and the item variables are conditionally independent given the properties. Moreover, we initially assume that each question $Q$ refers to a single property $C$ and that questions are conditionally independent given their relative property, as well as properties are conditionally independent given the item $I$. An example of this augmented setup is reported here below.

\begin{figure}[htp]
    \centering
    \begin{tikzpicture}[scale=.8]
    \node[nodo] (i)  at (3,1.5) {$I$};
    \node[nodo] (c1)  at (1,0) {\centering $C_1$};
    \node[nodo] (c2)  at (5,0) {\centering $C_2$};
    \node[Qnodo] (q1)  at (1.,-1.5) {$Q_1$};
    \node[Qnodo] (q2)  at (5,-1.5) {$Q_2$};
    \draw[arco] (c1) -- (q1);
    \draw[arco] (c2) -- (q2);
    \draw[arco] (i) -- (c1);
    \draw[arco] (i) -- (c2);
    \end{tikzpicture}
    \caption{Item, questions and properties relations}
    \label{fig:bn2}
\end{figure}
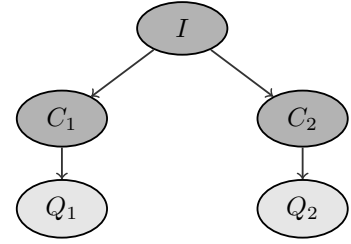

\begin{Example}
{\it In the same setup of Example \ref{ex:1}, consider two properties $C_1$ and $C_2$ modeling, respectively, the type of enterteiner and the type of event. Question $Q_1$ refers to property $C_1$, while $Q_2$ refers to $C_2$. Adopting the Markov condition for directed graphs \cite{koller2009}, the above discussed conditional independence assumptions correspond to the graph in Figure \ref{fig:bn2}}.
\end{Example}

The simplest way to define properties consists in regarding them as latent (i.e., not directly observable) clones of the questions, which are instead intended as manifest (i.e., observable) variables. In other words, if $Q$ is a question, we define a property $C$ whose possible states $\mathcal{C}$ are in one-to-one correspondence with the answers  $\mathcal{Q}$ and we set the logical constraint $C \equiv Q$ meaning that $P(Q=q|C=c)=1$ if $q$ and $c$ are the elements in the correspondence, and zero otherwise. In terms of compatibility functions, this means $\delta(i,c)=\delta(i,q)$ where $q$ and $c$ are the compatible states of $Q$ and $C$. Under the above assumptions, the marginalization of the properties simply returns a model equivalent to the one defined in the previous section. More expressive setups can be obtained by specifying properties that are not latent clones of the questions they are associated to. This allows to describe items based on general properties, streamlining the elicitation of compatibility relations between items and questions. For instance, a state $c$ of property $C$ can imply both logical requirements of the answers $q'$ and $q''$. Such a property would be compatible with both answers, but equivalent to none of them. This situation can be modeled by conditional probabilities $P(q'|c)$, $P(q''|c)$ strictly larger than zero. 
\begin{Example}
{\it In the Stagend platform, it is well known that the organizer of an event may be well satisfied with a band even if she asked only for a musician. This can be modeled by assigning a positive value to the probability $P(q_1^3|c_1^2)$ of asking for a musician when the best matching item $i^*$ is a band (i.e., $C(i^*) = c_1^2$ corresponding to \textit{Band}).}
\end{Example}

Moreover, the possibility of grouping items based on their properties can be exploited to learn model parameters with less fragmentation in the data. E.g., the probability $P(q|c)$ can be estimated from all selected items having property $c$; instead, to estimate $P(q|i)$ one should consider only the cases where item $i$ has been chosen, which can be very few or even zero for new items. 

Finally, more complex relations between questions and properties or property values can be easily modeled. Some examples are: (i) multiple questions referring to the same property, modeled by assuming conditional independence between the questions, this simply requires to independently elicit the probabilities $P(Q_1|C)$ and $P(Q_2|C)$ for all possible states of $Q_1$ and $Q_2$ and $C$; (ii) multiple properties for the same question, if $\mathbf{C}_Q$ is the set of properties associated with $Q$, the elicitation of the probabilities $P(q|\mathbf{C}_Q)$ for all values of $q$ and all joint combinations of states for the properties in $\mathbf{C}_Q$ is required; (iii) questions relevant only for some items, e.g., Stagend question \textit{Do you want the musician to play any particular instrument?} should define a preference among musicians, without changing the probability of the property \emph{musician} with respect to \emph{band}, \emph{DJ} and \emph{entertainer} and can be modeled by assuming $P(q|i) = P(q)$ for all items $i$ for which the question is not relevant.

\begin{Example} \label{ex:dependencies}
{\it In the Stagend example, the most requested entertainers for weddings are musicians and DJs. To model this we remove the assumption of marginal independence between the properties} Type of even {\it and} Type of entertainer{\it, i.e., $C_1$ and $C_2$. In the graphical language of Figure \ref{fig:bn2}, this corresponds to adding an edge connecting the two properties (Figure \ref{fig:bn3}). To define the corresponding model, the elicitation of the conditional probabilities $P(c_1|c_2,i)$ for all possible values of $C_1$ conditioned over all possible combinations of values $(c_2,i)$ is required. Then, in this toy example, where only three entertainers have been considered, adding the above dependency between properties requires elicitation of 36 parameters values which can become many more in real applications. It would be much easier to reason about the marginal probability $P(C_1|C_2)$ of the type of entertainer given the type of event.}

\begin{figure}[htp] \label{fig:bn3}
\centering
\begin{tikzpicture}[scale=.8]
\node[nodo] (i)  at (3,1.5) {$I$};
\node[nodo] (c1)  at (1,0) {\centering $C_1$};
\node[nodo] (c2)  at (5,0) {\centering $C_2$};
\node[Qnodo] (q1)  at (1.,-1.5) {$Q_1$};
\node[Qnodo] (q2)  at (5,-1.5) {$Q_2$};
\draw[arco] (c1) -- (q1);
\draw[arco] (c2) -- (q2);
\draw[arco] (i) -- (c1);
\draw[arco] (i) -- (c2);
\draw[arco] (c2) -- (c1);
\end{tikzpicture}
\caption{Graph representing the model in Example \ref{ex:dependencies}}
\end{figure}
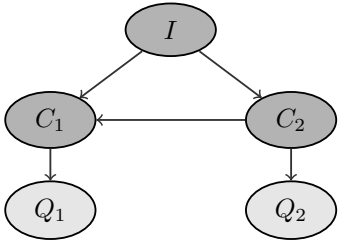
\end{Example}

\section{Elicitating Dependencies between Properties} 
In the previous section we noticed that adding dependencies between properties increases the number of model parameters (e.g., conditional probabilities), and this prevents a fast elicitation for huge catalogues. 

As we did for the conditional probability tables of the questions (i.e., $P(Q|C)$), we could better perform a single elicitation for all the items with the same value for the properties of interest, that is, focusing only on properties (and not on items) in the knowledge-based elicitation of probabilities, and grouping all items with the same property values when learning from data.

This can be achieved as follows. Assume we wish to model the conditional probability table for property $C$ given a set of parent properties  $\mathbf{C}_p$, while the remaining properties are assumed independent from $C$ and denoted by $\mathbf{C}_{np}$. The procedure for eliciting the (possibly huge) number of conditional probabilities $P(c|\mathbf{c}_p, i)$ for all values $c$ of $C$ and all joint combinations $(\mathbf{c}_p, i)$ consists of two steps. First, the conditional probabilities $P(c|\mathbf{c}_p)$ marginalized over $i$ are elicited based on prior knowledge, data or both, while the conditional probabilities $P(i|\mathbf{c})$, where $\mathbf{c} = [c, \mathbf{c}_p, \mathbf{c}_{np}]$ is the vector of all property values, is elicited based on logic constraints. Afterwards, probabilities $P(c|\mathbf{c}_p,i)$ are then be derived from the relation:
\begin{equation} \label{eq:ccpi}
P(c|\mathbf{c}_p,i) = \frac{P(i|\mathbf{c})P(c|\mathbf{c}_p)P(\mathbf{c}_p)}{\sum_{c}P(i|\mathbf{c})P(c|\mathbf{c}_p)P(\mathbf{c}_p)}\,,
\end{equation}
while the prior is derived from
\begin{equation}\label{eq:pi}
P(i) = \sum_{\mathbf{c}} P(i|\mathbf{c}) P(c|\mathbf{c}_p) P(\mathbf{c}_{np})\,.
\end{equation}
Concerning the conditional $P(i|\mathbf{c})$, it is derived from the joint $P(i,\mathbf{c})$ which has the same form as $P(i,\mathbf{q})$ in Equation \eqref{eq:genjoin} with $\mathbf{q}^J$ and $\mathbf{q}^P$ replaced by the vectors of properties $\mathbf{c}^J$ and $\mathbf{c}^P$ for which we assume, respectively, UJS and UPS. $P(i|\mathbf{c})$ is then given by: 
\begin{equation} \label{eq:ic}
P(i|\mathbf{c}) = \frac{\prod_{c_j \in \mathbf{c}} \delta(i,c_j)}{\sum_i\prod_{c_j \in \mathbf{c}} \delta(i,c_j)} = \frac{\delta(i,\mathbf{c})}{N
(\mathbf{c})}\,,
\end{equation}
where $N(\mathbf{c}) = \sum_i \delta(i,\mathbf{c})$ is the number of items compatible with the combination of property values in $\mathbf{c}$. 
Notice that in case no item is compatible with $\mathbf{c}$ the probability $P(i|\mathbf{c})$  cannot be derived from Equation \eqref{eq:ic}. 
This inconsistency arises from the fact that only the vectors of property values $\mathbf{c}$ that are compatible with at least one of the item in the catalogue, are, indeed, possible. Let $ \mathcal{J}_\mathbf{c}^{*}$ be the set of all possible $\mathbf{c}$ for which $\exists i: i \models \mathbf{c}$. To solve the above inconsistency and account for the logical impossibility of all $\mathbf{c} \not\in\mathcal{J}_\mathbf{c}^{*}$, an initial, eventually inconsistent, elicitation of empirical estimate $P'(\mathbf{c})$  of the probability of the joint properties needs to be revised as follows:
\begin{equation} \label{eq:pcrev}
P(\mathbf{c}) =
\left\{
\begin{array}{ll}
\frac{P'(\mathbf{c})}{\sum_{\mathbf{c} \in \mathcal{J}_\mathbf{c}^{*}} P'(\mathbf{c})}  & \text{if } \mathbf{c} \in \mathcal{J}_\mathbf{c}^{*}\,\\
0 & \text{otherwise}\,.
\end{array}
\right.
\end{equation}
Finally, by replacing Equation \eqref{eq:ic} into Equation \eqref{eq:ccpi} we obtain the conditional probabilities for the dependent property $C$:
\begin{equation} 
P(c|\mathbf{c}_p,i)= \frac{ \delta(i,c)\delta(i,\mathbf{c}_p) P(c|\mathbf{c}_p)}{\sum_{c^j: i\models c^j}  P(c|\mathbf{c}_p)}\,,
\end{equation}
which corresponds to the conditional $P(c|\mathbf{c}_p)$ re-normalized considering only the values of $c$ compatible with $i$.
By replacing  \ref{eq:ic} into \ref{eq:pi} we obtain the marginal distribution of $i$:
\begin{equation} 
P(i)= \sum_{\mathbf{c}: i\models \mathbf{c}} \frac{ P(\mathbf{c})}{N(\mathbf{c})}\,.
\end{equation}
Notice that the probability of each combination $\mathbf{c}$ of properties values is uniformly distributed over all items that are compatible with that combination. 

\begin{Example} \label{ex:invarrows}
{\it In the Stagend example the two steps above correspond to the elicitation of the conditional probabilities of the model in figure \ref{fig:bn4}. The only difference compared to the desired model (figure \ref{fig:bn3}) is the direction of the edges connecting $I$ to  $C_1$ and $C_2$. Therefore, the two models define the same set of independence assumptions. Assume that the relation between properties can be modeled by the conditional probabilities in the square brackets of Table \ref{tb:c1c2}, which describe a situation where the preferred types of entertainers are Djs and musicians for weddings, bands  and entertainers for corporate events, Djs and entertainers for birthdays, Djs and bands for parties. 
Moreover, we assume for $C_2$ the prior probabilities  $P(c_2^1) = 2/3$, $P(c_2^2) = P(c_2^3) = P(c_2^4) = 1/9$, according to which weddings are twice more popular than any other type of event. 
The corresponding joint model $P(C_1,C_2) = P(C_1|C_2)P(C_2)$ however, does not comply with the logical impossibility of all joint states $(c_1, c_2)$  that are not compatible with any item, requiring such joint states to have zero probability. As all values of $C_2$ are compatible with at least one item, we focus on $P(C_1|C_2)$. Cells corresponding to impossible joint states are highlighted in grey in Table \ref{tb:c1c2}. Their values are set to zero and the re-normalized probabilities are shown in the table next to the initial ones. 
 \begin{table}[ht] \label{tb:c1c2}
 \centering
 \begin{tabular}{ccccc}
 \hline
 $P(C_1|C_2)$ &$c^1_1$ {\tiny=DJ}&$c^2_1$ {\tiny=Band}&$c^3_1$ {\tiny=Musician}&$c^4_1$ {\tiny=Entert.}\\
 \hline
 $c^1_2$ {\tiny=Wedding}  & $\nicefrac{2}{3}$ ($\nicefrac{1}{3}$)& $\nicefrac{1}{3}$ ($\nicefrac{1}{6}$)  &  \cellcolor{black!30} $0$ ($\nicefrac{1}{3}$) & \cellcolor{black!30} $0$ ($\nicefrac{1}{6}$)\\
 $c^2_2$  {\tiny=Corp.Ev.} & $\nicefrac{1}{3}$ ($\nicefrac{1}{6}$) & $\nicefrac{2}{3}$ ($\nicefrac{1}{3}$) & \cellcolor{black!30} $0$ ($\nicefrac{1}{6}$) &  \cellcolor{black!30} $0$ ($\nicefrac{1}{3}$)\\
 $c^3_2$  {\tiny=Birthday}& $\nicefrac{1}{2}$ ($\nicefrac{1}{3}$)&  \cellcolor{black!30} $0$ ($\nicefrac{1}{6}$) &  \cellcolor{black!30} $0$ ($\nicefrac{1}{6}$)  & $\nicefrac{1}{2}$ ($\nicefrac{1}{3}$)\\
 $c^4_2$ {\tiny=Party} & $\nicefrac{2}{3}$ ($\nicefrac{1}{3}$)&   \cellcolor{black!30} $0$ ($\nicefrac{1}{3}$)&  \cellcolor{black!30} $0$ ($\nicefrac{1}{6}$)  & $\nicefrac{1}{3}$ ($\nicefrac{1}{6}$)\\
 \hline
 \end{tabular}
 \caption{Elicitation of $P(C_1|C_2)$}
 \end{table}

From this elicitation the prior probabilities $P(i_1) = 11/18$, $P(i_2) = 8/27$, $P(i_3) = 5/54$ and the conditional $P(C_1|C_2,i) = \delta(i,c_1)$ follow. Therefore, in this simple case where  each artist is compatible with one single  value of property $C_1$, we have that $P(C_1|C_2,i)$  reduces to the zero-entropy conditional $P(C_1|i)$ based only on logical constraints, whereas the defined dependence between $C_1$ and $C_2$ only affects the prior distribution over items. If a further entertainer $i_4$ performing at weddings both as DJs and as a band  were included in the catalog, it would result in strictly positive probabilities $P(c_1|c^1_2,i_4) = P(c_1|c^1_2)$ both for $c_1$ equal to band and Dj.}

\begin{figure}[htp!] \label{fig:bn4}
\centering
\begin{tikzpicture}[scale=.8]
\node[nodo] (i)  at (3,1.5) {$I$};
\node[nodo] (c1)  at (1,0) {\centering $C_1$};
\node[nodo] (c2)  at (5,0) {\centering $C_2$};
\node[Qnodo] (q1)  at (1.,-1.5) {$Q_1$};
\node[Qnodo] (q2)  at (5,-1.5) {$Q_2$};
\draw[arco] (c1) -- (q1);
\draw[arco] (c2) -- (q2);
\draw[arco] (c1) -- (i);
\draw[arco] (c2) -- (i);
\draw[arco] (c2) -- (c1);
\end{tikzpicture}
\caption{Graph representing the model in Example \ref{ex:invarrows}}
\end{figure}
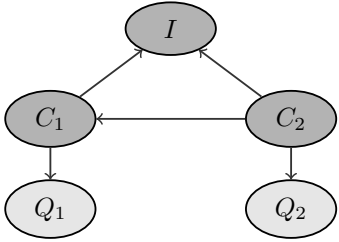
\end{Example}

\section{Learning from Data}
The above discussed procedure for the elicitation of the model parameters in the recommendation system is based on structural judgement about the logical compatibility between items, properties and the answers to be possibly integrated by judgements of a domain expert about the relations between the properties (irrespectively of the items). Yet, historical data involving observations of the parameters to be quantified might be available. Following a Bayesian paradigm, we can naturally merge these two sources of information by using the outputs of the elicitation process as the parameter of a multinomial Dirichlet prior to be combined with the likelihood of the observed data. This has the potential of further increasing the discriminative power of the system and the quality of the recommendations.

\section{Experiments}
For an empirical validation we consider the Stagend recommendation system. Currently, the platform includes $n=3520$ entertainers. In its static version, the system asks to all the interested users a questionnaire including $m=32$ questions intended to identify the entertainer that matches at best the needs of the users. Stagend advisors select a small subset of artists (in general less than ten) to be presented to each user. The questions identify $13$ properties used to characterize the entertainers. We model the relation between questions and items by means  of the compatibility-based elicitation and simulate a conversational process using the answers collected in $100$ questionnaires filled by actual Stagend users. At the end of each simulated conversation, a set of items is retained by our model. When all questions are asked, in $9$ cases over the $100$ simulated, the final set is empty. This shows that the notion of logical compatibility can be too strict in some cases. For the remaining questionnaires, we compare the decision of our algorithm to the subset of items selected by Stagend advisors. Let FI denote the fraction of items suggested by the advisors that are actually retained by our model, the average FI over the 91 remaining simulations is $70\%$. Again, this can be explained by the fact that logical compatibility is not always fully respected by the advisors suggestions, sometimes due to a poor knowledge of the entertainers in the catalogue, sometimes to better diversify the offer. 
\begin{figure}
\centering
\includegraphics[width=4.cm]{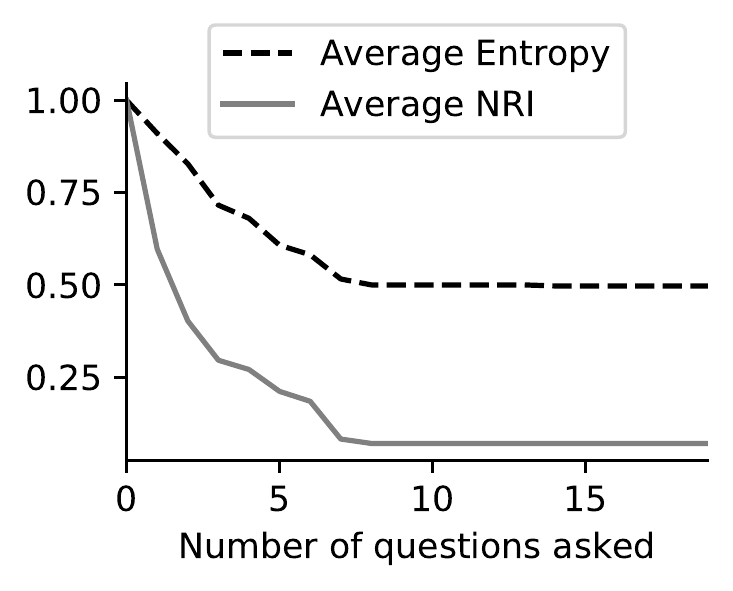}
\includegraphics[width=4.cm]{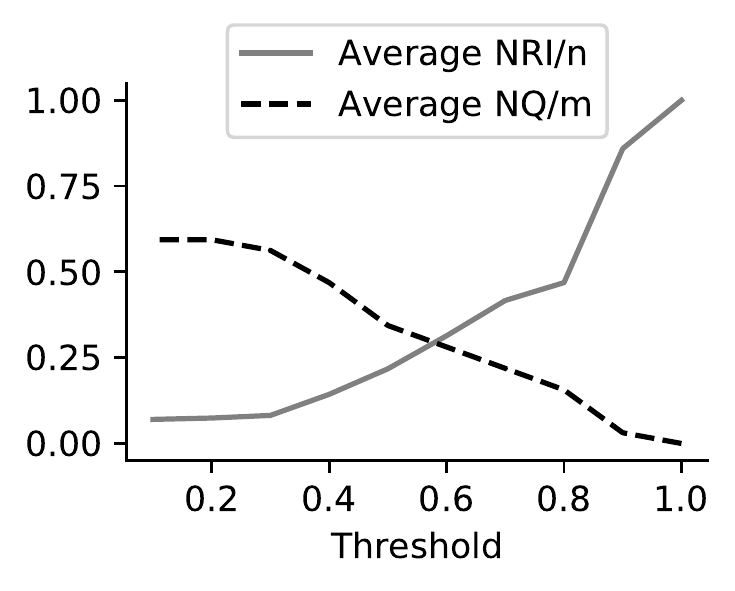}
\caption{Conversational approach performance} \label{adaptive}
\end{figure}
Figure \ref{adaptive} (left) shows how the entropy of $P(I|\mathbf{q})$ as well as the \emph{number of retained entertainers} (NRI) decreases with the number of question asked. However, after a certain number of questions, both the entropy and the NRI stop decreasing. Notice that the order of the question is defined in an adaptive way. In the right-hand side of Figure \ref{adaptive} it is shown how the fraction of retained entertainers  (NRI/$n$) and the fraction of questions asked (NQ/$m$) varies with respect to the entropy threshold.

\section{Conclusions}
A new approach to automatic recommendations which assumes a dynamic interaction between the system  and the user to provide customized and self-adaptive recommendations  has  been  developed  on  the  basis  of  a  pure Bayesian approach. The framework introduced in this paper sets the ground to several future developments, among which the dynamic generations of questions in order to improve the conversational nature of the system. This could be based on a natural language generation system interacting with the structured probabilistic description of item properties and elicitation of user needs, \cite{mostafazadeh2016generating}.
\bibliographystyle{aaai}
\bibliography{mangili}

\begin{thebibliography}{}

\bibitem[\protect\citeauthoryear{Adomavicius and
  Tuzhilin}{2011}]{adomavicius2011context}
Adomavicius, G., and Tuzhilin, A.
\newblock 2011.
\newblock Context-aware recommender systems.
\newblock In {\em Recommender systems handbook}. Springer.
\newblock  217--253.

\bibitem[\protect\citeauthoryear{Butz, Hua, and Maguire}{2006}]{butz2006web}
Butz, C.~J.; Hua, S.; and Maguire, R.~B.
\newblock 2006.
\newblock A web-based {B}ayesian intelligent tutoring system for computer
  programming.
\newblock {\em Web Intelligence and Agent Systems: An International Journal}
  4(1):77--97.

\bibitem[\protect\citeauthoryear{He, Parra, and
  Verbert}{2016}]{he2016interactive}
He, C.; Parra, D.; and Verbert, K.
\newblock 2016.
\newblock Interactive recommender systems: A survey of the state of the art and
  future research challenges and opportunities.
\newblock {\em Expert Systems with Applications} 56:9--27.

\bibitem[\protect\citeauthoryear{Jugovac and
  Jannach}{2017}]{jugovac2017interacting}
Jugovac, M., and Jannach, D.
\newblock 2017.
\newblock Interacting with recommenders - overview and research directions.
\newblock {\em ACM Transactions on Interactive Intelligent Systems} 7(3):10.

\bibitem[\protect\citeauthoryear{Koller and Friedman}{2009}]{koller2009}
Koller, D., and Friedman, N.
\newblock 2009.
\newblock {\em Probabilistic graphical models: principles and techniques}.
\newblock MIT press.

\bibitem[\protect\citeauthoryear{Lu \bgroup et al\mbox.\egroup
  }{2015}]{lu2015recommender}
Lu, J.; Wu, D.; Mao, M.; Wang, W.; and Zhang, G.
\newblock 2015.
\newblock Recommender system application developments: a survey.
\newblock {\em Decision Support Systems} 74:12--32.

\bibitem[\protect\citeauthoryear{Mangili, Bonesana, and
  Antonucci}{2017}]{mangili2017reliable}
Mangili, F.; Bonesana, C.; and Antonucci, A.
\newblock 2017.
\newblock Reliable knowledge-based adaptive tests by credal networks.
\newblock In {\em European Conference on Symbolic and Quantitative Approaches
  to Reasoning and Uncertainty},  282--291.
\newblock Springer.

\bibitem[\protect\citeauthoryear{Mill{\'a}n \bgroup et al\mbox.\egroup
  }{2000}]{Millan2002}
Mill{\'a}n, E.; Trella, M.; P{\'e}rez-de-la Cruz, J.-L.; and Conejo, R.
\newblock 2000.
\newblock Using {B}ayesian networks in computerized adaptive tests.
\newblock In {\em Computers and Education in the 21st Century}. Springer.
\newblock  217--228.

\bibitem[\protect\citeauthoryear{Mostafazadeh \bgroup et al\mbox.\egroup
  }{2016}]{mostafazadeh2016generating}
Mostafazadeh, N.; Misra, I.; Devlin, J.; Mitchell, M.; He, X.; and Vanderwende,
  L.
\newblock 2016.
\newblock Generating natural questions about an image.
\newblock {\em arXiv preprint arXiv:1603.06059}.

\bibitem[\protect\citeauthoryear{Sarwar \bgroup et al\mbox.\egroup
  }{2000}]{sarwar2000analysis}
Sarwar, B.; Karypis, G.; Konstan, J.; Riedl, J.; et~al.
\newblock 2000.
\newblock Analysis of recommendation algorithms for e-commerce.
\newblock In {\em EC},  158--167.

\bibitem[\protect\citeauthoryear{Wainer \bgroup et al\mbox.\egroup
  }{2000}]{wainer2000computerized}
Wainer, H.; Dorans, N.~J.; Flaugher, R.; Green, B.~F.; and Mislevy, R.~J.
\newblock 2000.
\newblock {\em Computerized adaptive testing: a primer}.
\newblock Routledge.

\end{thebibliography}
\end{document}